\documentclass[letter, 10 pt, conference]{ieeeconf}  % Comment this line out if you need a4paper

\IEEEoverridecommandlockouts                              % This command is only needed if 
\usepackage{url}
\usepackage{cite}
\usepackage{amsmath,amssymb,amsfonts}
\usepackage{algorithmic}
\usepackage{graphicx}
\usepackage{textcomp}
\usepackage{xcolor}
\usepackage{booktabs}
\usepackage{multirow}
\usepackage{cuted}
\usepackage{caption}
\usepackage{stfloats}
\usepackage{censor}

\usepackage{tablefootnote}

\def\BibTeX{{\rm B\kern-.05em{\sc i\kern-.025em b}\kern-.08em
    T\kern-.1667em\lower.7ex\hbox{E}\kern-.125emX}}
\begin{document}

\title{\LARGE \bf
MorphIK: Morphology-Conditioned Neural Inverse Kinematics for Unknown Robots
}

\author{
\authorblockN{Lennart Clasmeier$^1$}
%\authorblockA{University of Hamburg}
\and
\authorblockN{Jan-Gerrit Habekost$^1$}
%\authorblockA{University of Hamburg}
\and
\authorblockN{Cornelius Weber$^1$}
\and
\authorblockN{Stefan Wermter$^1$}
%\authorblockA{University of Hamburg}
\thanks{The authors gratefully acknowledge funding from Horizon Europe under the MSCA grant agreement No 101226624 (GREET) and support from the German Research Foundation DFG under project LUMO (No 551629603).}
\thanks{$^1$ L. Clasmeier, J.-G. Habekost, C. Weber and S. Wermter are with the Knowledge Technology Group, Department  of Informatics, University of Hamburg, Hamburg, Germany,
e-mail: \{lennart.clasmeier, jan-gerrit.habekost, stefan.wermter\}@uni-hamburg.de}
}

\maketitle
\begin{strip}
\vspace*{-7em}
    \centering
    \includegraphics[width=\linewidth]{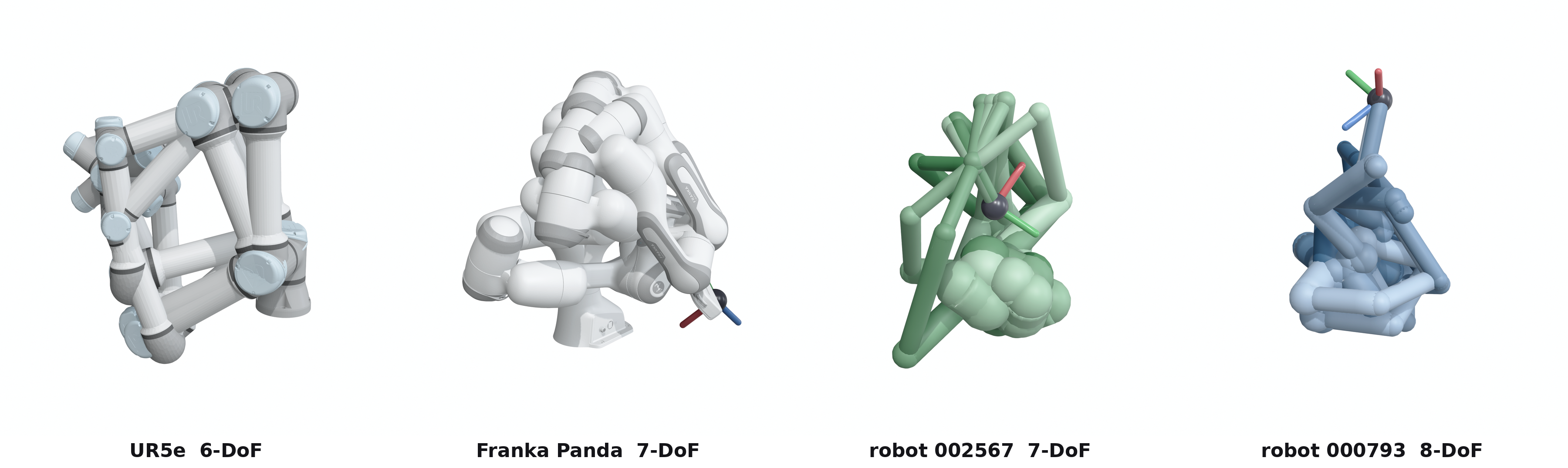}
    \captionof{figure}{Our neural inverse kinematics model generates a large variety of solutions for a single pose on real-world and procedurally generated arms, unseen during training} 
    \label{fig:nullspace}
\end{strip}

\begin{abstract}
Neural models can learn to generate various solutions to the inverse kinematics problem from data, but are usually limited to a single robot.
We present MorphIK, a flow-matching model that solves inverse kinematics for revolute-joint-based kinematic chains it has never seen during training.
The model uses a transformer architecture to encode the robot's morphology along with the target pose.
This encoding then conditions a flow-matching head that generates poses from noise.
Trained on purely synthetic data from procedurally generated robots, the model reaches a precision of about 5 cm on unseen real-world robots with 6 to 9 Degrees of Freedom.
For higher precision, the model serves as an excellent Prior for further optimization algorithms, reducing error to less than 1 cm after a single step of Damped Least Squares optimization and to sub-1 mm error after 3 steps in most cases.
Building on flow matching's generative capabilities to produce highly diverse outputs, our model can efficiently sample the robot's null space, providing a wide variety of configurations for the same pose.
Thus, overall MorphIK allows learning and generalizing neural inverse kinematics for a multitude of known and unknown robots. 
\end{abstract}


\section{Introduction}
Although inverse kinematics is ubiquitous in robotics, it remains an open challenge for redundant kinematic chains, with many approaches in parallel for different use cases.
One family of approaches that has gained traction in recent years uses neural methods to learn an inverse kinematics model from data, rather than treating it as an optimization problem or seeking a closed-form solution.
This approach offers multiple advantages over traditional methods, including runtime, redundancy, and singularity handling.
Since neural inference can run batched on a GPU, many poses can be solved at once, enabling parallel exploration of the null space.
Modern generative approaches like normalizing flows\cite{Ames_2022}, GANs\cite{Habekost_2024}, diffusion\cite{zhang2026ikdiffuserdiffusionbasedgenerativeinverse}, or flow matching\cite{yang2026mimicik} model one-to-many solutions and therefore are a natural fit for the multi-modality of Inverse Kinematics for redundant manipulators.
Learned models also do not rely on the kinematic chain's Jacobian, so singular configurations become less severe.

While these advantages are promising, the biggest limitation of neural IK solvers is that they are usually single-robot models trained to solve the IK problem for exactly one kinematic chain.
This results in costly data collection and retraining, a real hurdle for integrating and adapting these approaches.
We propose MorphIK, a transformer-based neural multi-robot IK solver that generalizes to all kinds of 6- to 9-DoF unseen revolute serial manipulators.
Our model requires only a robot's URDF, from which we derive a canonical representation and tokenize it as the model's geometry input.

A Transformer backbone combined with a flow-matching decoder generates IK solutions for a given pose target.
Because the robot is an input rather than a fixed assumption, training requires not one robot but many. Thus, we train on a multitude of randomly generated robots covering a wide range of morphologies.
We generate training data in a fully self-supervised way by sampling configurations from the robot's joint spaces and applying forward kinematics to obtain the corresponding poses.
We store only robot descriptions and randomly generate poses on the GPU during training.
Although the model has never seen a real robot during training, we show that it generalizes to a wide range of real-world manipulators, achieving less than 1 cm error after a single step and sub-millimeter accuracy after only 3 steps of post-inference polishing with DLS.

\section{Related Work}
Many different IK algorithms have been proposed over the years.
Traditional solvers like KDL\cite{kdl-url} or Trac-IK \cite{Beeson_2015} use the Jacobian of the kinematic chain to solve the problem, which can cause issues near singularities where the Jacobian determinant is 0 and velocities become infinite.
Although damping and singularity avoidance can address these issues, they still create real limitations that must be worked around. 

Genetic algorithms do not rely on the Jacobian, which makes singularities less of an issue for algorithms like Bio-IK \cite{Starke_2020}.
Additionally, this class of algorithms can adapt to arbitrary constraints added to the optimization objective, but at the cost of higher computational overhead.

Neural Methods try to overcome this issue by learning to solve Inverse Kinematics from data.
The main advantage is that inference on modern GPUs is extremely fast and parallel, and generative architectures can model multimodal distributions, allowing effective exploration of the robot's null space.
Further, different loss functions can adapt the model behavior to the desired behavior, and training data can be generated by calculating forward kinematics on random joint configurations.
%cheap data through sampling in joint space and fk
%loss can be done in cart space through differentiable FK

IK Flow \cite{Ames_2022} uses normalizing flows to train a generative IK network to an accuracy of 1 cm and $2^\circ$ of error.
Additionally, it serves as a seed model for more effective null space sampling in Trac-IK.
CycleIK \cite{Habekost_2024} trains both a GAN and an MLP model for IK by utilizing differentiable forward kinematics to calculate position and orientation loss in Cartesian space.
Generative Graphical Inverse Kinematics \cite{Limoyo_2025} trains a Graph Neural Network to solve the Inverse Kinematics problem by means of completing a partial graph of distances.
It can incorporate multiple robots in the same model through a distance-geometric graph representation of manipulators using a set of fixed points.
It also trains a model on a randomly generated dataset of robots, then evaluates it on real-world robotic manipulators with error ranges similar to our approach.
IKDiffuser\cite{zhang2026ikdiffuserdiffusionbasedgenerativeinverse} generalizes neural IK to kinematic trees with multiple end effectors.
By tokenizing the end effector together with a target, a single model can solve IK for multiple chains of the same robot.
While the model provides a single formula applicable to all kinds of morphologies, such as hand kinematic trees, it remains a single-robot model that must be trained individually for each robot.
MimicIK\cite{yang2026mimicik} uses flow matching to train a neural IK solver for teleoperation, focusing on trajectory smoothness and real-time performance from teleoperation data.
We extend the flow matching approach to a larger conditioning vector, which includes a robot representation, turning it into a multi-robot model.

\section{Approach}
Our approach uses a tokenized canonical robot representation as conditioning for a flow matching model that predicts robot configurations.
Since this conditioning requires a large variety of robots to generalize, we created a dataset of procedurally generated robots for training.
We add a Cartesian auxiliary loss to the flow matching training objective, which speeds up and stabilizes training.
Finally, we evaluate the model on a set of unseen real-world robots.
\begin{table*}[b]
\caption{Dataset metrics for procedurally generated and real-world robots at median, [5th percentile, and 95th percentile]}
\centering
\begin{tabular}{lrrrrrrrr}
\toprule
 & robots & joint range (rad) & longest link (m) & manipulability $\times 10^{-3}$ \cite{yoshikawa1985manipulability} & condition number\cite{angeles1992kinematic} & 3 parallel & 3 meeting \\
\midrule
6-DoF & 99800 & 4.05 [2.03, 6.06] & 0.37 [0.26, 0.56] & 1.84 [0.37, 7.00] & 71.81 [33.76, 171.44] & 5\% & 38\% \\
7-DoF & 99800 & 3.90 [1.49, 6.05] & 0.36 [0.25, 0.56] & 6.34 [1.16, 21.85] & 42.92 [22.26, 122.00] & 9\% & 64\%\\
8-DoF & 99800 & 3.91 [1.59, 6.04] & 0.33 [0.23, 0.53] & 13.15 [2.96, 37.93] & 32.65 [19.31, 83.28] & 12\% & 76\%  \\
9-DoF & 99800 & 3.92 [1.71, 6.05] & 0.31 [0.21, 0.49] & 22.18 [6.15, 56.76] & 27.95 [18.08, 61.16] & 13\% & 85\% \\  \addlinespace
%all & 399200 & 3.94 [1.82, 6.05] & 0.34 [0.23, 0.54] & 8.57 [0.74, 40.28] & 39.24 [19.97, 130.70] & 10\% & 66\%\\
real 6-DoF & 14 & 6.28 [2.89, 6.28] & 0.40 [0.31, 0.47] & 8.92 [3.78, 19.44] & 31.72 [20.04, 59.16] & 57\% & 29\%\\
real 7-DoF & 8 & 5.87 [3.64, 6.28] & 0.38 [0.35, 0.45] & 30.22 [23.62, 74.12] & 19.17 [15.50, 23.04] & 12\% & 75\% \\
real 8-DoF & 14 & 5.04 [0.70, 6.28] & 0.34 [0.28, 0.48] & 53.24 [37.34, 123.59] & 15.21 [11.95, 26.66] & 54\% & 62\%  \\
real 9-DoF & 8 & 5.75 [0.70, 6.28] & 0.34 [0.28, 0.44] & 135.29 [59.27, 210.97] & 11.94 [10.23, 14.23] & 14\% & 86\%  \\
\bottomrule
\end{tabular}
\label{tab:dataset}
\end{table*}

\begin{figure}
    \vspace{5pt}
    \centering
    \includegraphics[width=1\linewidth]{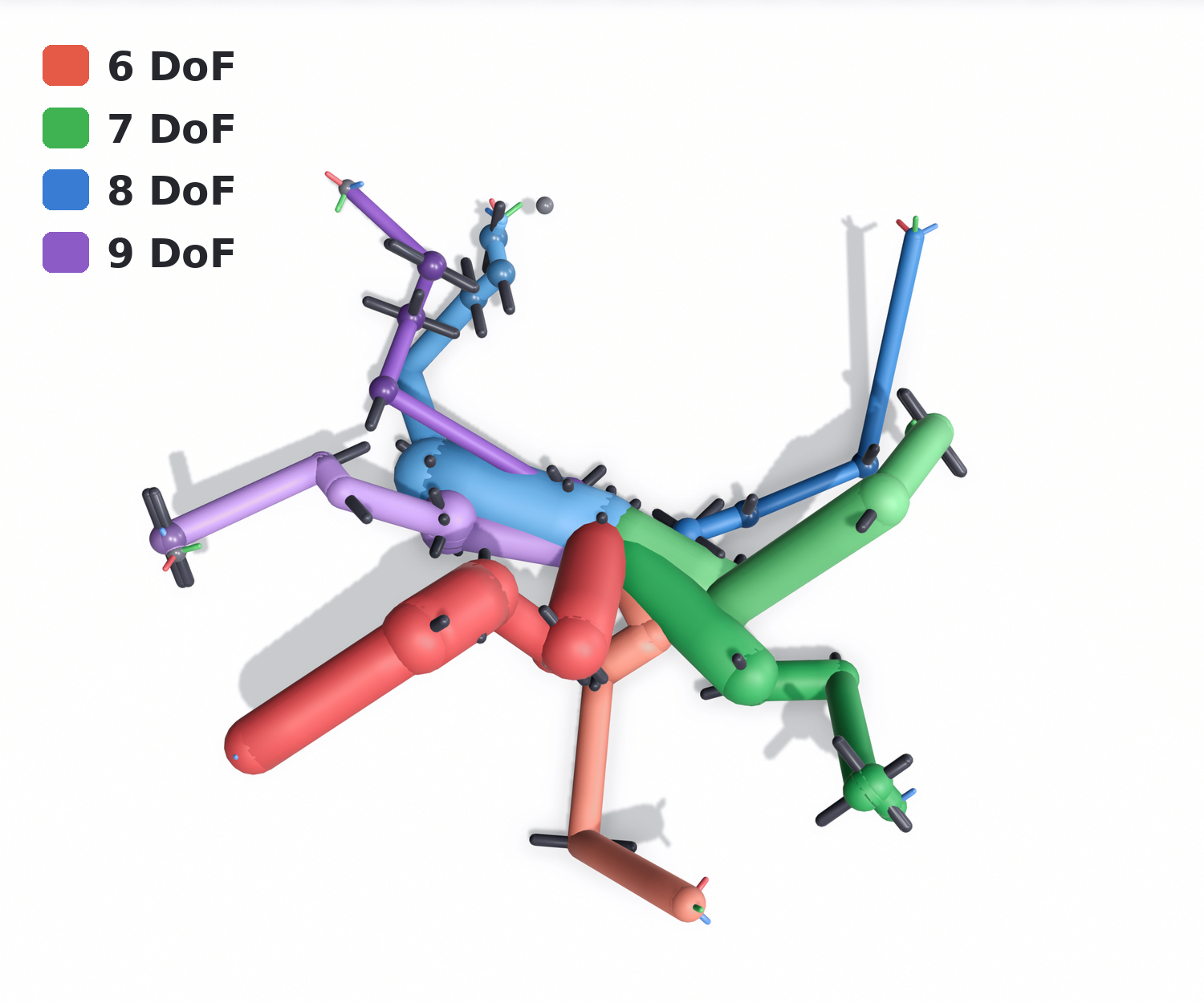}
    \caption{Example robots from the procedural robot generator}
    \label{fig:robot_zoo}
\end{figure}

\subsection{Canonical Robot Representation}
Our approach generalizes neural IK to unseen robots by using procedurally generated robots as data generators.
The generator samples random revolute kinematic chains and stores them in a canonical form.
The canonical form removes redundancies from the robot representation, making robot morphology representations easier to learn because identical robots have identical representations.
The robot is described as a chain of joints, each defined by a rotation axis in space through an axis point and a unit vector for the axis direction.
Since we can choose the axis point freely by sliding along the axis, we canonicalize it as the point closest to the previous axis point in the chain.
The first joint defines the origin: its axis becomes the +z axis, and the axis point is placed where that axis comes closest to the next axis that is not parallel to it.
To remove the remaining rotational freedom around the z-axis, we rotate about $z$ until the first link that leaves the axis lies in the $xz$ plane.
Each remaining axis direction is fixed by requiring its first non-negligible component to be positive; flipping an axis mirrors that joint's rotation, so its limits are negated and swapped.
One freedom survives both steps: reversing the first joint's axis describes the same robot. 
By construction, we rotate that axis onto +z before we can decide its direction.
We therefore canonicalize both variants and use the parameter vector with the smaller value at the first differing entry as a tie-breaker.
A rigid tool transform describing the end-effector pose completes the robot description.
Finally, we normalize the total chain length, including the tool segment, to one to create scale invariance.
The result is a product-of-exponentials description: each joint is a screw axis expressed in one common frame, and forward kinematics is the product of the corresponding matrix exponentials\cite{brockett2005robotic}.

Because all joints share one reference frame rather than the relative frames of the URDF or DH conventions, the resulting geometry tokens are directly comparable across joints, robots, and degrees of freedom, and the model never has to reconstruct the kinematic chain to relate them.

\subsection{Dataset}
We sample joint links from a Dirichlet distribution ($\alpha = 2$), with 35\% of the links set to zero, which makes the adjacent axes intersect and creates pseudo-spherical joints. The rotational axes are sampled in a structured way: 25\% of the joints are parallel to the previous joint, 55\% orthogonal to it, and 20\% uniformly on the sphere.
This mix creates a highly varied set of synthetic kinematic chains to help with generalization, while introducing some structure resembling real-world robots.

For 7-, 8-, and 9-DoF chains, roughly 30\% of the robots terminate in a compact two-joint wrist rather than distributing joints evenly along the whole chain.
This layout is inspired by real-world scenarios, in which a robotic hand with its own wrist is attached to a cobot.

Joint limits are symmetric around zero, with the half-range drawn uniformly from $[0.9, \pi]$ for arm joints and $[0.15, 0.9]$ for wrist joints, since wrists are usually more limited in their movement. Placing the 0 value at the center of the range is a convention rather than a property of the mechanism, and the URDF importer re-centers every real arm the same way.
Tool orientation is sampled uniformly from SO(3) as a unit quaternion.
The Capsule radii are sampled with a slight taper towards the end of the chain.

We sample 100{,}000 robots per DoF class, leading to a dataset of 400{,}000 robots, holding out 200 robots per class for evaluation.
During training, we generate data online from this fixed set of robots, providing an unlimited supply of training samples.
First, we sample joint values from the bounds, then check configurations for self-collision using a simple capsule model.
Finally, a GPU-accelerated, highly parallelized FK solver computes the forward kinematics, producing the final training triplet of robot description, target pose, and joint configuration.
To compare the model's ability to generalize to real hardware, we also evaluate it on a set of 6- and 7-DoF robotic arms from the robot\_descriptions\cite{robotdescriptions_py} Python package as real-world robot geometries.
We further generated realistic 8- and 9-DoF chains by attaching the Shadow Hand, also available in this package and featuring a 3-DoF wrist, to the 6- and 7-DoF robots and deactivating its redundant wrist rotation joint.
This yields 44 real robot geometries, in addition to the 800 held-out synthetic robots from the robot generator.

Table \ref{tab:dataset} shows the recorded metrics of the procedurally generated robot multitude compared to the evaluated 6-, 7-, 8-, and 9-DoF real-world kinematic chains. 
Our robots match the real arms in link length (about 35 cm), but their joint ranges are distributed differently, and their kinematic properties differ more substantially.
While real manipulators often have large joint ranges that allow near-full motion, with some very limited joints like wrists, we sampled the synthetic joint ranges uniformly.
We report manipulability \cite{yoshikawa1985manipulability} as a measure of how far a configuration sits from a singularity.
It is the volume of the velocity ellipsoid at a single configuration, so larger values mean more end-effector motion is available per unit of joint motion.
Since it is defined pointwise, we report the median over 32 randomly sampled configurations per robot.
We also report the condition number of the Jacobian \cite{angeles1992kinematic}, the ratio of the ellipsoid's longest to shortest semi-axis, which describes the shape rather than the size of the same ellipsoid; smaller is better, with 1 denoting an arm equally capable in every direction.

On both measures, the generated robots are significantly worse than the hand-crafted ones, and consistently so across DoF classes: their manipulability is roughly five times lower and their condition numbers roughly twice as high.
In practice, this means the generated arms spend far more of their configuration space near singularities, and their motion capability is much more unevenly distributed across directions.
Both properties affect the inverse kinematics directly, since a Jacobian-based solver must invert exactly this matrix at every iteration: a condition number of 72 rather than 32 means errors are amplified more than twice as much in the linear solve that the polishing step performs.
The generated robots are therefore not simply unfamiliar but numerically harder to solve than the real arms it is evaluated on.

Another common evaluation metric for robotic manipulators is the Pieper criterion \cite{pieper1969kinematics}, which defines whether a 6-DoF kinematic chain has a closed-form solution.
A closed-form solution exists either if three consecutive joints are parallel, or if three consecutive axes meet in a single point, forming a spherical wrist.
While 86\% of the real 6-DoF arms satisfy at least one of the two Pieper criteria, only 42\% of the generated 6-DoF chains do. The gap is entirely in the parallel case: 57\% of real 6-DoF arms carry three consecutive parallel axes against 5\% of ours, whereas on the intersecting criterion the generated chains slightly exceed the real ones (38\% against 29\%). The share rises with DoF for both populations because more joints yield more consecutive triples that may satisfy either criterion.

\subsection{Model Architecture}
\begin{figure*}
    \vspace{5pt}
    \centering
    \includegraphics[width=1\linewidth]{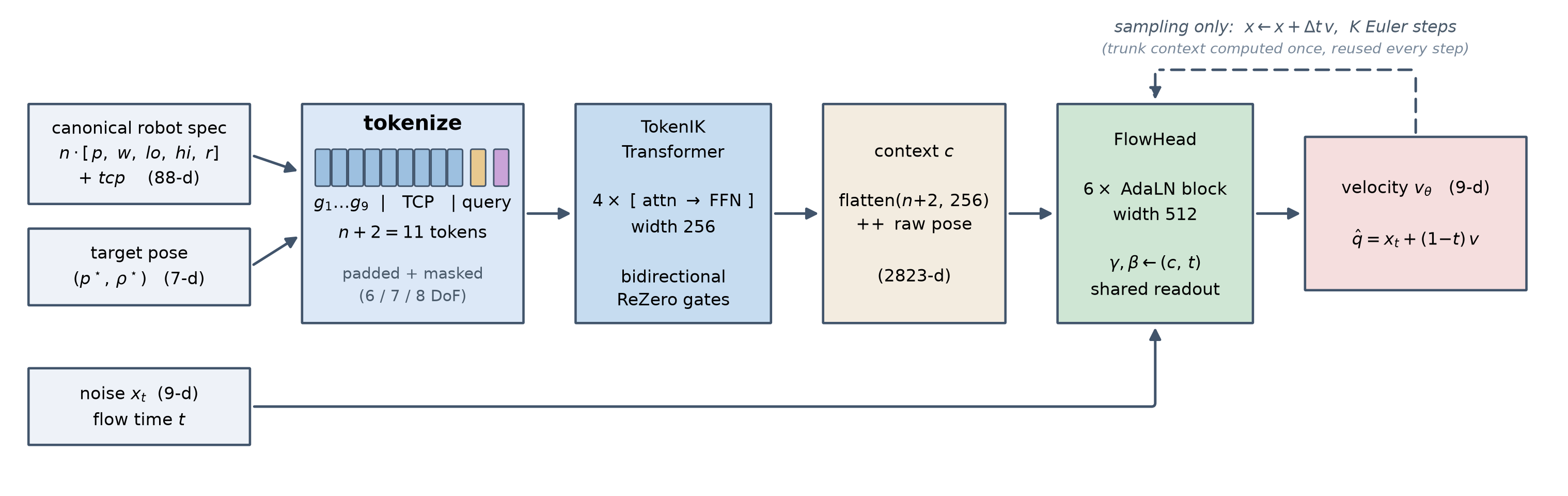}
    \caption{Robot Geometry Tokens and Target pose are tokenized and used as input to the Transformer backbone. The model output is then flattened and used to condition the flow Matching Head. The flow head takes noise and flow time $ t$ as input and generates the velocity vector $v_\theta$. At inference, this vector is applied to $x_t$ and fed back into the flow head to create the next step.}
    \label{fig:model_arch}
\end{figure*}

\begin{figure}
    \centering
    \includegraphics[width=\linewidth]{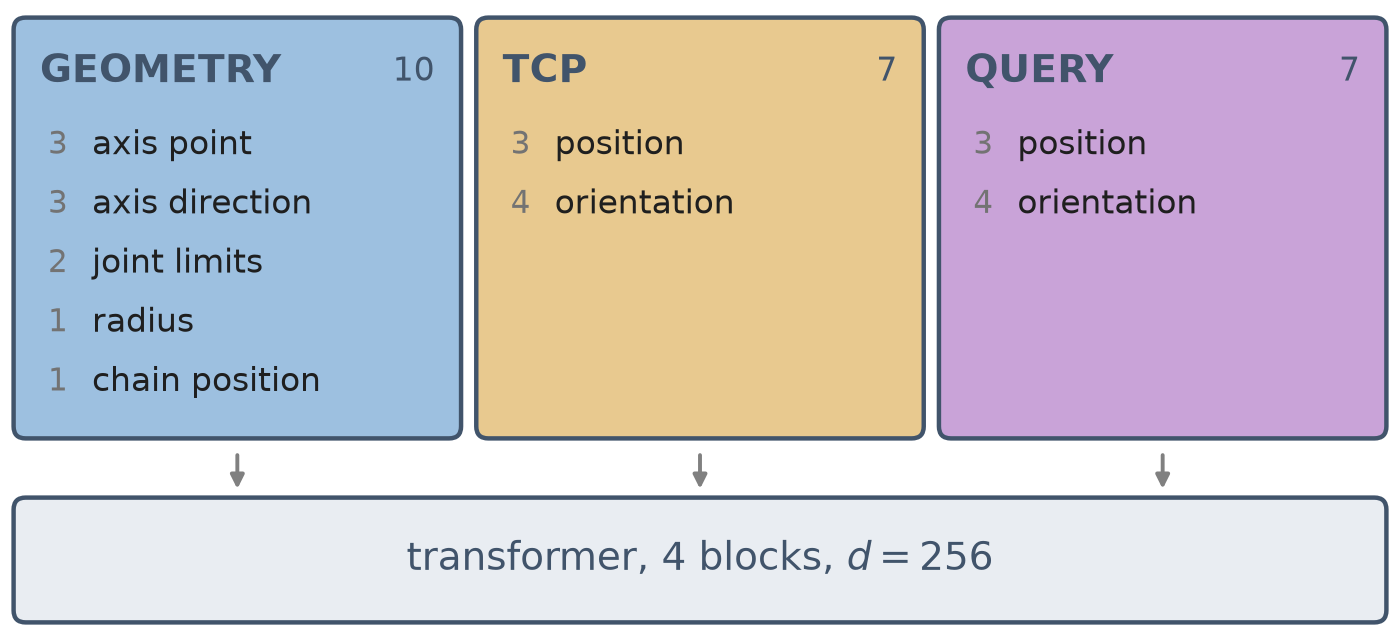}
    \caption{Token layout of the MorphIK transformer backbone. The Geometry tokens contain the robot's morphological information, with one token per joint. A single TCP token represents the end-effector pose at joint configuration 0, and a Query token defines the IK Pose target.}
    \label{fig:tokens}
\end{figure}

The neural architecture shown in Fig.~\ref {fig:model_arch} uses a transformer trunk to encode the robot geometry and task, and a flow-matching head to generate the robot configuration.
Similar to the MetaMorph architecture\cite{gupta2022metamorph}, the encoder takes a sequence of geometry tokens containing one token per joint and a tool center point (TCP) token, plus a pose token that encodes the target end-effector pose as shown in Fig~\ref{fig:tokens}.
While MetaMorph only handles robots from its own recipe, our canonical robot representation ensures that all kinds of robots can be encoded directly from URDF into the same space.
The encoder generates the encoding with multiple rounds of multi-headed self-attention in a traditional encoder-only transformer architecture.
The encoder output of $n+2$ token embeddings is flattened into a conditioning vector for the flow matching head \cite{lipman2023flow}.

The flow matching head is an MLP with AdaLN \cite{peebles2023scalable} conditioning. The flow matching head takes noise as input and acts as a velocity field that integrates the noise over multiple passes.
The AdaLN layer takes the task encoding from the transformer backbone as input.
All robot architecture token sequences are padded and masked for the maximum 9-DoF sequence length, similar to the approach of $\pi_0$\cite{black2024pi0}. 

\subsection{Losses}
The model is trained on a combination of flow matching loss and forward kinematic loss.
Equation \ref{eq:LossFM} shows the standard flow matching loss, with $x_0$ being a random vector, and $q$ the joint configuration.
$ v^*$ is the target velocity from the noise vector $x_0$ to the goal configuration $q$, and $\mathcal{L}_{\mathrm{FM}}$ is the MSE between the predicted velocity and the target velocity $v^*$

\begin{align}
    x_t &= (1-t)\,x_0 + t\,q,
      \qquad v^{*} = q - x_0 \\[3pt]
\mathcal{L}_{\mathrm{FM}} &= \bigl\| v_\theta(x_t,\,t,\,c) - v^{*} \bigr\|^{2} \label{eq:LossFM}
\end{align}
 
Habekost et al. \cite{Habekost_2024} proposed a cycle loss enabled by differentiable forward kinematics to compute position and orientation loss in Cartesian space, providing a strong training signal for IK and allowing deterministic networks to commit to a single solution rather than averaging over multiple configurations for the same pose.
We bring this approach into the domain of flow matching as an auxiliary loss that heavily stabilizes the training.
Normal diffusion approaches struggle with this kind of loss because it requires an actual final output $q$.
Since the learned vector field is trained without full inference, only relatively noisy estimates are available \cite{ho2020denoising}.

The flow matching model's velocity field is mostly straight, so we can estimate the trajectory endpoint in a single step by following the predicted velocity directly to the end, yielding an estimate $\hat{q}$ of the output configuration.
We then use this estimated output configuration to compute the estimated end-effector pose $(\hat{p}, \hat{q})$ via differentiable forward kinematics $\mathrm{FK}$ and, from that, the Cartesian position and orientation losses $\mathcal{L_{\mathrm{ori}}}$ and $\mathcal{L_{\mathrm{pos}}}$.

\begin{align}
    \hat{q} &= x_t + (1-t)\,v_\theta(x_t,\,t,\,c) \\[3pt] \label{eq:one_step_q}
(\hat{p},\,\hat{o}) &= \mathrm{FK}(\hat{q}) \\[3pt]
\mathcal{L}_{\mathrm{pos}} &= \mathrm{smooth}_{L_1}\bigl(\hat{p},\; p^{*}\bigr) \\[3pt]
\mathcal{L}_{\mathrm{ori}} &= \min\bigl\{\, \mathrm{smooth}_{L_1}(\hat{o},\; o^{*}),
    \mathrm{smooth}_{L_1}(-\hat{o},\; o^{*}) \,\bigr\}
\end{align}

Since the estimated joint position is only useful at the end of integration, and predictions tend to be poor at the beginning of the flow process, we time-gate the forward kinematic losses to the second half of the flow process.
Additionally, the losses are scaled by $t^2$ which results in the time gate factor $\gamma(t)$

\begin{align}
\gamma(t) &= \begin{cases}
               t^{2}, & t \ge \tfrac{1}{2}\\[2pt]
               0,     & \text{otherwise}
             \end{cases} 
\end{align}

The final loss $\mathcal{L}$ is the expected value of the weighted sum of the three loss terms.

\begin{align}
\mathcal{L} &= \mathbb{E}\Bigl[\,
      \lambda_{\mathrm{FM}}\,\mathcal{L}_{\mathrm{FM}}
    + \lambda_{\mathrm{FK}}\,\gamma(t)\bigl(
        \omega_p\,\mathcal{L}_{\mathrm{pos}}
      + \omega_o\,\mathcal{L}_{\mathrm{ori}}\bigr) \Bigr]
\end{align}

The transformer encoder and the flow matching decoder are trained end-to-end through this combined training objective.

\subsection{Training}
We trained the model for 1{,}000{,}000 steps with a batch size of 2048.
We kept the learning rate constant at 5.4e-4 for the first 350{,}000 steps, then linearly annealed it to 8e-6.
The forward kinematic loss weight $\lambda_{\mathrm{FK}}$ was weighted 6 to 1 to the flow matching weight $\lambda_{\mathrm{FM}}$.
For the FK loss, the position weight factor $\omega_p$ was weighted 5 to 1 relative to the orientation weight $\omega_o$.
We used the model's exponential moving average (EMA) weights to reduce noise, as it is common to flow matching models.
We evaluated the model with 10 flow-matching steps.
For the experiments that only require a single solution, we inferred the model with a batch size of 16 and selected the best solution for further DLS optimization and reporting.
All experiments were done on a single Nvidia RTX 5080 with 16GB of VRAM.

\section{Experiments}
In our experiments, we evaluate three different applications of the proposed model.
We evaluate it as a raw IK solver, as a fine-tuned Foundation model for a single robot, as well as a seed model for other optimization approaches like DLS\cite{wampler1986manipulator}.
To show the model's generalization capabilities, we evaluate it on two groups of robots: held-out robots, unseen during training but created by the same generator, and real-world robots, which were unseen during training.
For the fine-tuning experiments, we randomly picked three real-world robots from each DoF and fine-tuned the model only on these samples.
To show the model's capabilities as a foundation model, we also trained the same model from scratch on these robots and report its performance after the same number of steps.

\begin{table*}[]
\vspace{5pt}
\caption{Performance of the different robot groups (best of 16): raw MorphIK result and after 1, 3, and 5 DLS steps}
\centering
  \begin{tabular}{ll|rr|rr|rr|rr}
\toprule
 &  & \multicolumn{2}{c}{raw} & \multicolumn{2}{c}{+1 DLS} & \multicolumn{2}{c}{+3 DLS} & \multicolumn{2}{c}{+5 DLS} \\
\cmidrule(lr){3-4}\cmidrule(lr){5-6}\cmidrule(lr){7-8}\cmidrule(lr){9-10}
 &  & \scriptsize{pos} & \scriptsize{ori} & \scriptsize{pos} & \scriptsize{ori} & \scriptsize{pos} & \scriptsize{ori} & \scriptsize{pos} & \scriptsize{ori} \\
\midrule
\multicolumn{10}{l}{\scriptsize\emph{position in \% of reach, orientation in deg}} \\
\multirow{3}{*}{Heldout 6-DoF \scriptsize{(200)}} & \scriptsize{mean} & 3.434 & 2.72 & 0.737 & 0.41 & 0.221 & 0.09 & 0.145 & 0.08 \\
 & \scriptsize{median} & 3.126 & 2.45 & 0.474 & 0.20 & 0.087 & 0.01 & 0.040 & 0.00 \\
 & \scriptsize{p99} & 9.034 & 7.50 & 4.870 & 3.87 & 2.476 & 1.50 & 1.981 & 1.21 \\
\midrule
\multirow{3}{*}{Heldout 7-DoF \scriptsize{(200)}} & \scriptsize{mean} & 3.446 & 2.65 & 0.625 & 0.41 & 0.149 & 0.08 & 0.089 & 0.07 \\
 & \scriptsize{median} & 3.150 & 2.39 & 0.408 & 0.19 & 0.053 & 0.00 & 0.020 & 0.00 \\
 & \scriptsize{p99} & 8.968 & 7.22 & 3.979 & 3.93 & 1.780 & 1.63 & 1.256 & 1.41 \\
\midrule
\multirow{3}{*}{Heldout 8-DoF \scriptsize{(200)}} & \scriptsize{mean} & 3.704 & 2.79 & 0.579 & 0.42 & 0.101 & 0.06 & 0.051 & 0.04 \\
 & \scriptsize{median} & 3.424 & 2.55 & 0.396 & 0.21 & 0.034 & 0.00 & 0.010 & 0.00 \\
 & \scriptsize{p99} & 9.306 & 7.37 & 3.291 & 3.66 & 1.161 & 1.29 & 0.731 & 0.94 \\
\midrule
\multirow{3}{*}{Heldout 9-DoF \scriptsize{(200)}} & \scriptsize{mean} & 4.102 & 3.21 & 0.606 & 0.49 & 0.074 & 0.04 & 0.030 & 0.03 \\
 & \scriptsize{median} & 3.805 & 2.95 & 0.428 & 0.27 & 0.026 & 0.00 & 0.006 & 0.00 \\
 & \scriptsize{p99} & 10.124 & 8.23 & 3.158 & 3.73 & 0.812 & 0.84 & 0.431 & 0.53 \\
\midrule
\multirow{3}{*}{\textbf{Heldout, all} \scriptsize{(800)}} & \scriptsize{mean} & 3.671 & 2.84 & 0.637 & 0.43 & 0.136 & 0.07 & 0.079 & 0.05 \\
 & \scriptsize{median} & 3.376 & 2.58 & 0.427 & 0.22 & 0.050 & 0.00 & 0.019 & 0.00 \\
 & \scriptsize{p99} & 9.358 & 7.58 & 3.824 & 3.80 & 1.557 & 1.31 & 1.100 & 1.02 \\
\midrule
\multirow{3}{*}{Real 6-DoF \scriptsize{(14)}} & \scriptsize{mean} & 4.216 & 3.34 & 0.599 & 0.39 & 0.078 & 0.02 & 0.041 & 0.01 \\
 & \scriptsize{median} & 3.904 & 3.07 & 0.344 & 0.17 & 0.017 & 0.00 & 0.003 & 0.00 \\
 & \scriptsize{p99} & 10.519 & 8.77 & 4.263 & 3.59 & 0.767 & 0.21 & 0.486 & 0.08 \\
\midrule
\multirow{3}{*}{Real 7-DoF \scriptsize{(8)}} & \scriptsize{mean} & 3.999 & 3.40 & 0.323 & 0.26 & 0.016 & 0.00 & 0.007 & 0.00 \\
 & \scriptsize{median} & 3.754 & 3.21 & 0.221 & 0.17 & 0.001 & 0.00 & 0.000 & 0.00 \\
 & \scriptsize{p99} & 9.722 & 7.96 & 1.644 & 1.73 & 0.223 & 0.05 & 0.109 & 0.04 \\
\midrule
\multirow{3}{*}{Real 8-DoF \scriptsize{(14)}} & \scriptsize{mean} & 3.936 & 2.95 & 0.297 & 0.24 & 0.013 & 0.01 & 0.005 & 0.01 \\
 & \scriptsize{median} & 3.646 & 2.70 & 0.195 & 0.13 & 0.003 & 0.00 & 0.000 & 0.00 \\
 & \scriptsize{p99} & 9.561 & 7.40 & 1.649 & 1.76 & 0.148 & 0.07 & 0.061 & 0.05 \\
\midrule
\multirow{3}{*}{Real 9-DoF \scriptsize{(8)}} & \scriptsize{mean} & 3.965 & 3.11 & 0.205 & 0.17 & 0.005 & 0.00 & 0.001 & 0.00 \\
 & \scriptsize{median} & 3.725 & 2.95 & 0.154 & 0.12 & 0.000 & 0.00 & 0.000 & 0.00 \\
 & \scriptsize{p99} & 9.369 & 7.14 & 0.884 & 0.95 & 0.076 & 0.03 & 0.026 & 0.03 \\
\midrule
\multirow{3}{*}{\textbf{Real, all} \scriptsize{(44)}} & \scriptsize{mean} & 4.042 & 3.18 & 0.381 & 0.28 & 0.033 & 0.01 & 0.016 & 0.01 \\
 & \scriptsize{median} & 3.762 & 2.95 & 0.240 & 0.15 & 0.006 & 0.00 & 0.001 & 0.00 \\
 & \scriptsize{p99} & 9.860 & 7.89 & 2.341 & 2.19 & 0.345 & 0.10 & 0.199 & 0.05 \\
\midrule
\multirow{3}{*}{\textbf{Heldout, all (random init)} \scriptsize{(800)}} & \scriptsize{mean} & 42.087 & 57.02 & 29.044 & 43.47 & 12.266 & 19.95 & 5.027 & 7.72 \\
 & \scriptsize{median} & 40.098 & 55.89 & 25.851 & 42.00 & 9.104 & 17.53 & 2.414 & 3.59 \\
 & \scriptsize{p99} & 90.046 & 110.21 & 76.893 & 92.06 & 50.447 & 61.05 & 31.511 & 41.34 \\
\midrule
\multirow{3}{*}{\textbf{Real, all (random init)} \scriptsize{(44)}} & \scriptsize{mean} & 53.105 & 58.93 & 35.966 & 44.89 & 13.754 & 21.09 & 4.476 & 7.80 \\
 & \scriptsize{median} & 50.999 & 57.38 & 32.416 & 43.31 & 9.656 & 18.89 & 1.776 & 3.97 \\
 & \scriptsize{p99} & 114.611 & 113.66 & 94.657 & 95.21 & 58.522 & 62.54 & 31.968 & 40.69 \\
\bottomrule
\end{tabular}

\label{tab:Performance}
\end{table*}

\section{Results}
Table \ref{tab:Performance} shows the model's performance on the different groups of evaluated robots.
Since flow matching lets us sample multiple solutions for the same pose, we sample 16 configurations per pose and report the best, utilizing the flow matching model's parallel inference and providing a realistic setup that prevents model failure.
On average, synthetic robots reach a raw mean position error of 3.67\% of reach and an orientation error of about $2.84^{\circ}$.
The tail of the error distribution is about 2.5 times worse, with about 9.36\% position and $7.58^\circ$ orientation error.
On the synthetic robots, 6- and 7-DoF robots are very close, with the higher-DoF robots performing slightly worse.
On the real-world robots, all DoF perform similarly, with the 6-DoF robots performing slightly worse.

A single step of Damped Least Squares (DLS) optimization reduces the mean position error by a factor of 4 to below the 1\% range and brings the mean orientation error to less than $1^\circ$.
Optimizing the model output with DLS has a stronger effect for higher-DoF robots, reducing the mean and median error faster.
Real-world 6-DoF robots keep a 0.6\% reach error after one step, while the same robots with a wrist show about half that error.
Polishing makes the error distribution more tail-heavy. 3 DLS steps are generally enough to bring the median error into the 0.01\% range, while the 99th percentile stays high.
This is especially true for the 6-DoF robots, which still have a 1.98\% position error at the 99th percentile after 5 DLS steps.

Although the evaluated real-world robots are not sampled from the same distribution as the training data and are hand-designed by humans, the model's solutions for these robots tend to be better than those for the random robots.

While the raw model output is comparable to the one for the random robots, the hand-crafted architectures show much larger gains from DLS optimization.
A single step already brings the mean error of the real-world robots down to about half a centimeter, and 3 steps lead to sub-millimeter precision.
This behavior is basically mirrored for the orientation.

\begin{figure}
\vspace{5pt}
\centering
    \includegraphics[width=1\linewidth]{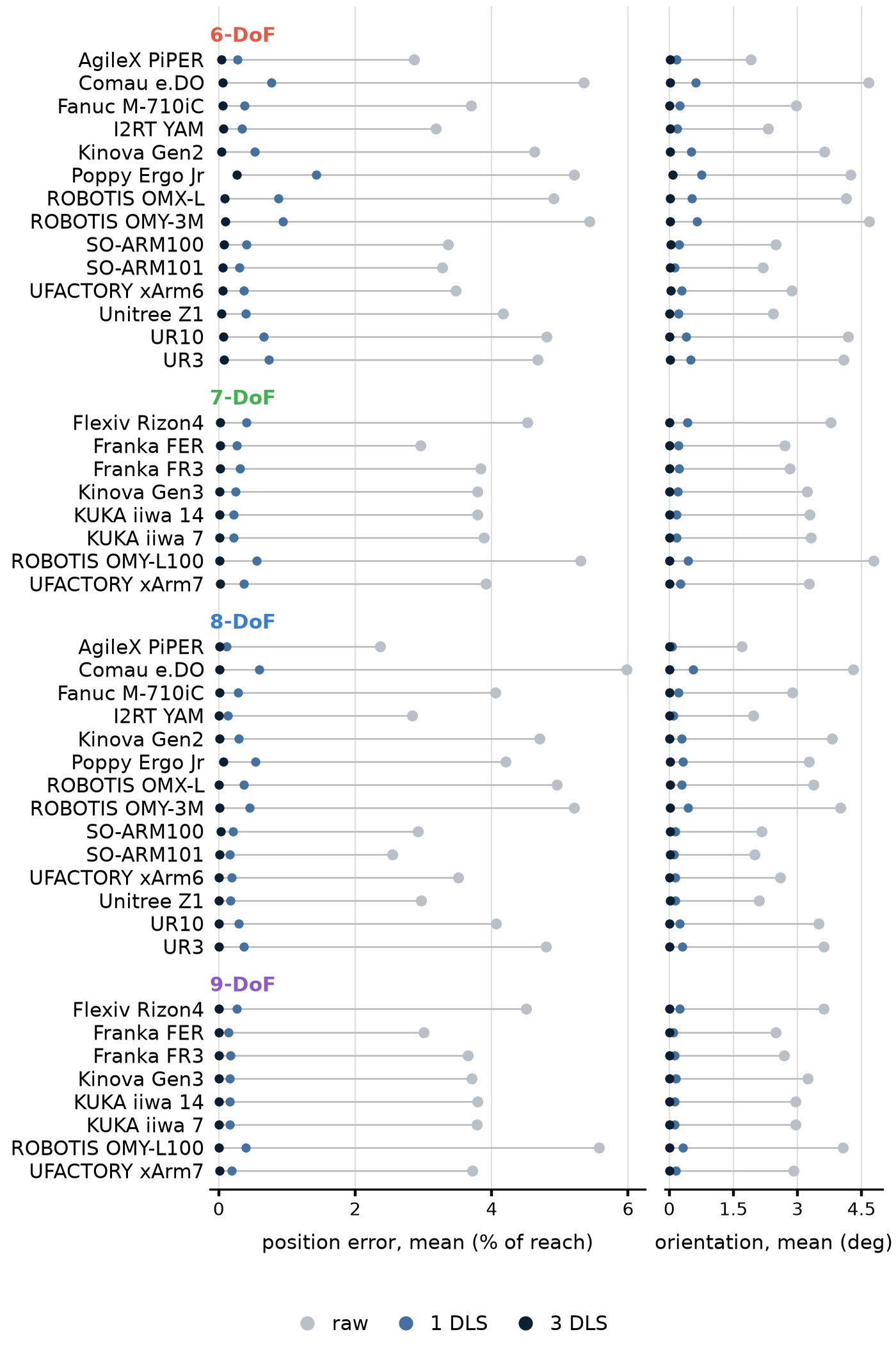}
    \caption{Mean performance of unseen real-world robots on 512 random poses. Raw and after 1 and 3  steps.} 
    \label{fig:real_robots}
\end{figure}

Figure \ref{fig:real_robots} shows the real robots' performance on a per-robot level.
For a fair comparison, we scaled the robots to unit length, since robot chain length strongly affects the position error.
Position-wise, the piper with the shadow hand performs best with 2.37\%, and the Comaue e.DO with the shadow-hand wrist performs worst at 5.98\% of reach.
Orientation-wise, the best is again the piper at $1.69^\circ$, and the worst is ROBOTIS OMY-L100 at $4.78^\circ$.
Overall, the arms perform very similarly with and without the attached wrist, as shown by the similarity between results 6- and 8-DoF and between 7- and 9-DoF. This suggests that most error comes from the arm, which makes sense since early joints in the chain have a stronger impact on overall position error than joints at the end of the chain.
While the wrist does not seem to affect raw error much, the same is not true for the DLS-optimized outputs.
When we compare the same arms with and without a wrist after 1 DLS step, the 8- and 9-DoF variants optimize faster than their 6- and 7-DoF counterparts.

\subsection{Null space Modeling}
Table \ref{tab:nullspace} reports the spread of generated solutions for the same target pose.
We sampled 256 random poses for each robot. For each pose, we generated 128 solutions and optimized them in 3 DLS steps.
The spec shows the counts of solutions that land within 1\% and $2^\circ$.
We report variety measures only for in-spec solutions, since randomly failed solutions would be distinct and increase the standard deviation but carry no information about the model's null space exploration.
$distinct_{0.25}$ is the number of solutions that differ by at least 0.25 radians in one joint compared to all other solutions; std[rad] reports the standard deviation of the in-spec solutions.

6-DoF Manipulators are a special case in this evaluation, since they do not have a null space but a fixed set of up to 16 distinct redundant configurations for a given pose, as shown by \cite{raghavan1993inverse}.
For the synthetic robots, about half of the samples per pose meet the 1 cm and $2^\circ$ spec in 3 DLS steps. 
For 6-DoF, we found a mean of 3.4 different configurations per pose and a standard deviation of 0.28 rad.
Absolute distinct counts as well as standard deviation increase with more degrees of freedom, which is expected and a general sign of a higher-dimensional null space of the higher-DoF chains.

The real robots show slightly different and better behavior.
For the 6-DoF arms, the model finds 6.2 configurations on average.
The number increases to 42.8 for the 7-DoF arms, which have a well-conditioned null space they can use.
We also see an increase in in-spec solutions, which we suspect results from better manipulator conditioning that makes DLS more effective.
The 8-DoF arms behave slightly worse than the set of 7-DoF arms. 
Since these were constructed by adding a wrist with smaller joint ranges to a 6-DoF arm, they are more limited in motion than the 7-DoF arms, although they have one more degree of freedom as shown in Table \ref {tab:dataset}.
Adding a wrist to the 7-DoF arm greatly improves redundancy, yielding an average of 80.5 distinct configurations out of 96.1 in spec solutions.
At this point, almost every solution is distinct, so the count is bounded by the number of samples instead of the model's capabilities.

\begin{table}[]
\vspace{5pt}
\centering
\caption{Redundancy Spread for the same pose}
\label{tab:nullspace}
\begin{tabular}{l|rr|r}
\toprule
 & \scriptsize{in spec / 128} & \scriptsize{distinct$_{0.25}$} & \scriptsize{std [rad]} \\
\midrule
\multicolumn{4}{l}{\scriptsize\emph{heldout synthetic robots, 128 samples per pose, 3 DLS steps}} \\
Heldout 6-DoF \scriptsize{(200)} & 61.8 & 3.4 & 0.28 \\
Heldout 7-DoF \scriptsize{(200)} & 66.1 & 11.9 & 0.49 \\
Heldout 8-DoF \scriptsize{(200)} & 68.1 & 32.4 & 0.66 \\
Heldout 9-DoF \scriptsize{(200)} & 69.0 & 55.3 & 0.81 \\
\midrule
\multicolumn{4}{l}{\scriptsize\emph{real arms from, 128 samples per pose, 3 DLS steps}} \\
Real 6-DoF \scriptsize{(14)} & 77.0 & 6.2 & 0.76 \\
Real 7-DoF \scriptsize{(8)} & 89.5 & 42.8 & 1.14 \\
Real 8-DoF \scriptsize{(14)} & 84.3 & 34.0 & 0.88 \\
Real 9-DoF \scriptsize{(8)} & 96.1 & 80.5 & 1.16 \\
\bottomrule
\end{tabular}

\end{table}

\begin{table}[b]
\caption{Fine-tuning vs from scratch training}
\centering
\begin{tabular}{rr|rr|rr}
\toprule
 &  & \multicolumn{2}{c}{pretrained} & \multicolumn{2}{c}{from scratch}\\
\cmidrule(lr){3-4}\cmidrule(lr){5-6}
\scriptsize{steps} & \scriptsize{time} & \scriptsize{pos [\%]} & \scriptsize{ori [$^\circ$]} & \scriptsize{pos [\%]} & \scriptsize{ori [$^\circ$]} \\
\midrule
0 & -- & 5.58 & 4.65 & 52.49 & 57.62\\
50 & 1\,s & 3.91 & 3.55 & 52.14 & 57.77 \\
300 & 4\,s & 3.11 & 3.09 & 39.72 & 53.66 \\
1000 & 14\,s & 2.62 & 2.89 & 18.66 & 40.04\\
3000 & 42\,s & 2.17 & 2.53 & 11.80 & 30.18 \\
20000 & 5\,min & 1.48 & 1.85 & 4.41 & 9.35 \\
\bottomrule
\end{tabular}

\label{tab:fine-tuning}
\end{table}
\subsection{Finetuning}
As shown above, our model generalizes well over a large variety of robot morphologies and provides a strong prior for post-inference optimization through DLS.
As a second possible application, we propose using the model as an IK Foundation model that can be fine-tuned for a specific robot.
While this approach reduces the model's ability to generalize, we expect it to improve raw output performance on a given robot while keeping the training budget low compared to training from scratch.
To test this capability, we randomly selected 8 real robots and trained the model only on these (one model per robot).
In another validation setting, we also trained the model for these robots from scratch for the same number of steps.

Table \ref{tab:fine-tuning} shows the results of this experiment.
Since these are aggregated numbers across multiple robots, we report the position error as a percentage of reach rather than in mm.
The experiment shows that even a very small amount of fine-tuning can drastically reduce the error.
After 50 steps, which corresponds to 1 second on the used hardware, the position error drops by 1.6\% of reach and $1.1^\circ$, corresponding to about a 30\% improvement in position and 25\% in orientation.
In that short time frame, the from-scratch model learns nothing.
After 1000 steps, the position error is halved compared to the baseline model, while the from-scratch model still fails on orientation and slowly learns position.
Generally, the gap between the pretrained and the from-scratch control shrinks with longer training. After 20{,}000, the gap reduced to a factor of 3, with the pretrained model reaching a mean positional error of 1.48\% of reach and the control reaching 4.41\%.
This is expected, since training gains slow over time and the control starts catching up to the fine-tuned model, although it does not reach it within the given time frame.
This experiment shows that fine-tuning the model to a single robot can significantly improve performance, even with a tight budget.
Compared to the from-scratch control group, it shows that the general model can serve as a strong starting point for training unseen robots.
While it does not unlock capabilities that a from-scratch model could not learn, it gives any robot in this experiment a strong head start.

\subsection{Time}
Table \ref{tab:time} shows the model's inference time per pose.
DLS optimization and best-config selection add a constant time cost of 5 ms and 1.3 ms, respectively.
DLS scales linearly with iterations at about 1.7 ms per step and is independent of batch size.
Model inference is variable with the number of inferred poses.
Inferring a single pose with a best-of-16 approach costs about 3.5 ms.
This per-pose inference cost drops drastically when we infer multiple poses per batch: 8 poses cost 0.5 ms per pose, and batches with 512 elements cost about 0.16 ms per pose. 
The constant upfront cost of DLS and pose selection, plus relative speed gains from batched inference, make the approach ideal for sampling large batches of poses at once, e.g., for planning.

\begin{table}[]
\vspace{5pt}
\centering
\caption{Inference Time for different batch sizes}
\begin{tabular}{r|cccc|cc}
\toprule
\scriptsize{poses} & \scriptsize{model} &  \scriptsize{m. per pose} &\scriptsize{+3 DLS} & \scriptsize{select} & \scriptsize{total} & \scriptsize{per pose} \\
 & \scriptsize{[ms]} & \scriptsize{[ms]} & \scriptsize{[ms]} & \scriptsize{[ms]} & \scriptsize{[ms]} \\
\midrule
1 & 3.5 & 3.5 & 5.0 & 1.3 & 9.8 & 9.85 \\
8 & 4.1 & 0.5 & 5.1 & 1.3 & 10.4 & 1.30 \\
64 & 13.5 & 0.2 &  5.1 & 1.3 & 19.9 & 0.31 \\
512 & 82.7 & 0.16& 5.1 & 1.3 & 89.1 & 0.17 \\
\bottomrule
\end{tabular}

\label{tab:time}
\end{table}

\section{Conclusion}
MorphIK provides a new multi-robot architecture for robot-independent training.
Trained only on procedurally generated robotic arms, \mbox{MorphIK} generalizes to a wide range of real 6- to 9-DoF robotic arms by learning to embed the morphological characteristics of kinematic chains.
The proposed multi-robot architecture performs strongly across a wide range of robot morphologies, achieving a mean positional error of about 4\% of the robots' reach and a $3.2^\circ$ orientation error without any further setup.
Further, we showed that the generated joint configurations serve as excellent priors for further optimization approaches like DLS, reducing the residual error in 3 steps to sub-0.1\% of reach and $1^\circ$ range, which corresponds to a sub-1 mm position error on regular-sized arms, making the model viable and attractive even for high-precision tasks.
One further main advantage of our model over traditional non-neural generative methods is its ability to explore the manipulators' null space in parallel.
In future work, we will examine the learned embedding space for different tasks and explore a transformer-based decoder to make the model truly DoF-independent.

\section{Acknowledgment}
The authors produced and validated all scientific content. AI-based tools, including Grammarly and Claude Opus 5 (Anthropic), were used to develop auxiliary code and check grammar.

\bibliographystyle{IEEEtran}
\bibliography{bib}

@article{Limoyo_2025,
   title={Generative Graphical Inverse Kinematics},
   volume={41},
   ISSN={1941-0468},
   DOI={10.1109/tro.2024.3521862},
   journal={IEEE Transactions on Robotics},
   publisher={Institute of Electrical and Electronics Engineers (IEEE)},
   author={Limoyo, Oliver and Marić, Filip and Giamou, Matthew and Alexson, Petra and Petrović, Ivan and Kelly, Jonathan},
   year={2025},
   pages={1002–1018} }

@article{Ames_2022,
  author       = {Barrett Ames and
                  Jeremy Morgan and
                  George Konidaris},
  title        = {{IKFlow}: Generating Diverse Inverse Kinematics Solutions},
  journal      = {CoRR},
  volume       = {abs/2111.08933},
  year         = {2021},
  eprinttype   = {arXiv},
  eprint       = {2111.08933},
  bibsource    = {dblp computer science bibliography, https://dblp.org}
}

@INPROCEEDINGS{Habekost_2024,
  author={Habekost, Jan-Gerrit and Gäde, Connor and Allgeuer, Philipp and Wermter, Stefan},
  booktitle={2024 IEEE/RSJ International Conference on Intelligent Robots and Systems (IROS)}, 
  title={Inverse Kinematics for Neuro-Robotic Grasping with Humanoid Embodied Agents}, 
  year={2024},
  volume={},
  number={},
  pages={7315-7322},
  doi={10.1109/IROS58592.2024.10802010}}

@phdthesis{Starke_2020,
  title={{Bio IK}: A memetic evolutionary algorithm for generic multi-objective inverse kinematics},
  author={Starke, Sebastian},
  year={2020},
  school={Staats-und Universit{\"a}tsbibliothek Hamburg Carl von Ossietzky}
}

@INPROCEEDINGS{Beeson_2015,
  author={Beeson, Patrick and Ames, Barrett},
  booktitle={2015 IEEE-RAS 15th International Conference on Humanoid Robots (Humanoids)}, 
  title={{TRAC-IK}: An open-source library for improved solving of generic inverse kinematics}, 
  year={2015},
  volume={},
  number={},
  pages={928-935},
  doi={10.1109/HUMANOIDS.2015.7363472}}

@Misc{            kdl-url,
  author        = {Smits, R.},
  title         = {{KDL}: {K}inematics and {D}ynamics {L}ibrary},
  howpublished  = {\url{http://www.orocos.org/kdl}},
}

@inproceedings{
lipman2023flow,
title={Flow Matching for Generative Modeling},
author={Yaron Lipman and Ricky T. Q. Chen and Heli Ben-Hamu and Maximilian Nickel and Matthew Le},
booktitle={The Eleventh International Conference on Learning Representations },
year={2023},
}

@inproceedings{
gupta2022metamorph,
title={MetaMorph: Learning Universal Controllers with Transformers},
author={Agrim Gupta and Linxi Fan and Surya Ganguli and Li Fei-Fei},
booktitle={International Conference on Learning Representations},
year={2022},
}

@inproceedings{peebles2023scalable,
  title={Scalable diffusion models with transformers},
  author={Peebles, William \& Xie, Saining},
  booktitle={2023 IEEE/CVF International Conference on Computer Vision (ICCV)},
  pages={4172--4182},
  year={2023},
  organization={IEEE}
}

@article{black2024pi0,
  title={$\pi_0 $: A Vision-Language-Action Flow Model for General Robot Control},
  author={Black, Kevin and Brown, Noah and Driess, Danny and Esmail, Adnan and Equi, Michael and Finn, Chelsea and Fusai, Niccolo and Groom, Lachy and Hausman, Karol and Ichter, Brian and others},
  journal={arXiv preprint arXiv:2410.24164},
  year={2024}
}

@article{yoshikawa1985manipulability,
  title={Manipulability of robotic mechanisms},
  author={Yoshikawa, Tsuneo},
  journal={The International Journal of Robotics Research},
  volume={4},
  number={2},
  pages={3--9},
  year={1985},
  publisher={Sage Publications Sage CA: Thousand Oaks, CA}
}

@book{pieper1969kinematics,
  title={The kinematics of manipulators under computer control},
  author={Pieper, Donald Lee},
  year={1969},
  publisher={Stanford University}
}

@inproceedings{brockett2005robotic,
  title={Robotic manipulators and the product of exponentials formula},
  author={Brockett, Roger W},
  booktitle={Mathematical Theory of Networks and Systems: Proceedings of the MTNS-83 International Symposium Beer Sheva, Israel, June 20--24, 1983},
  pages={120--129},
  year={2005},
  organization={Springer}
}

@article{wampler1986manipulator,
  title={Manipulator inverse kinematic solutions based on vector formulations and damped least-squares methods},
  author={Wampler, Charles W},
  journal={IEEE Transactions on Systems, Man, and Cybernetics},
  volume={16},
  number={1},
  pages={93--101},
  year={1986},
  publisher={IEEE}
}

@article{ho2020denoising,
  title={Denoising diffusion probabilistic models},
  author={Ho, Jonathan and Jain, Ajay and Abbeel, Pieter},
  journal={Advances in Neural Information Processing Systems},
  volume={33},
  pages={6840--6851},
  year={2020}
}

@misc{zhang2026ikdiffuserdiffusionbasedgenerativeinverse,
      title={{IKDiffuser}: a Diffusion-based Generative Inverse Kinematics Solver for Kinematic Trees}, 
      author={Zeyu Zhang and Ziyuan Jiao},
      year={2026},
      eprint={2506.13087},
      archivePrefix={arXiv},
      primaryClass={cs.RO},
}

@article{angeles1992kinematic,
  title={Kinematic isotropy and the conditioning index of serial robotic manipulators},
  author={Angeles, Jorge and L{\'o}pez-Caj{\'u}n, Carlos S},
  journal={The International Journal of Robotics Research},
  volume={11},
  number={6},
  pages={560--571},
  year={1992},
  publisher={Sage Publications Sage CA: Thousand Oaks, CA}
}

@article{raghavan1993inverse,
  title={Inverse kinematics of the general 6R manipulator and related linkages},
  author={Raghavan, Madhusudan and Roth, Bernard},
  year={1993}
}

@software{robotdescriptions_py,
  title = {{robot\_descriptions.py: Robot descriptions in Python}},
  author = {Caron, Stéphane and Romualdi, Giulio and Kozlov, Lev and Ordoñez Apraez, Daniel Felipe and Tadashi Kussaba, Hugo and Bang, Seung Hyeon and Zakka, Kevin and Schramm, Fabian and Uru\c{c}, Jafar and Traversaro, Silvio and Zamora, Jonathan and Castro, Sebastian and Tao, Haixuan Xavier and Yu, Justin and Jallet, Wilson and Zhang, Yutong and Walker, Nick and Bragin, Nikita and Wie, Woojin},
  license = {Apache-2.0},
  version = {3.1.0},
  year = {2026}
}

@article{yang2026mimicik,
  title={{MimicIK}: Real-Time Generative Inverse Kinematics from Teleoperation with FK Consistency},
  author={Yang, Jiahao and Yan, Shenhao and Feng, Fan and Yao, Chengsi and Wang, Ge and Mai, Zhixin and Zhao, Yiming and Han, Yatong},
  journal={arXiv preprint arXiv:2606.15148},
  year={2026}
}

\end{document}